\documentclass[conference]{IEEEtran}
\IEEEoverridecommandlockouts
\usepackage{cite}
\usepackage{amsmath,amssymb,amsfonts}
\usepackage{algorithmic}
\usepackage{graphicx}
\usepackage{subfig}
\usepackage{algorithm}
\usepackage{textcomp}
\usepackage{xcolor}
\usepackage[hidelinks]{hyperref}
\hypersetup{
    colorlinks,
    linkcolor={blue!70!black},
    citecolor={blue!70!black},
    urlcolor={white!10!black}
}

\def\gS{{\mathcal{S}}}
\newcommand{\E}{\mathbb{E}}
\def\gG{{\mathcal{G}}}
\def\eps{{\epsilon}}
\def\gD{{\mathcal{D}}}
\def\gB{{\mathcal{B}}}

\newcommand{\envname}{\textbf{LANRO}}
\newcommand{\heir}{\textbf{HEIR}}
\newcommand{\hipss}{\textbf{HIPSS}}
\newcommand{\lcsac}{\textbf{LCSAC}}
\newcommand{\say}[1]{\textit{``#1''}}
\newcommand{\sourcealgo}{https://github.com/knowledgetechnologyuhh/hipss}
\newcommand{\sourceenv}{https://github.com/knowledgetechnologyuhh/lanro-gym}

\newcommand{\mdefault}{\textit{Default}}
\newcommand{\mcolor}{\textit{Color}}
\newcommand{\mshape}{\textit{Shape}}
\newcommand{\mcolorshape}{\textit{ColorShape}}

\newcommand{\policy}{\pi}
\newcommand{\discount}{\gamma}

\usepackage{graphicx,calc}
\newlength\myheight
\newlength\mydepth
\settototalheight\myheight{Xygp}
\newcommand*\inlinegraphics[1]{%
  \settototalheight\myheight{Xygp}%
  \settodepth\mydepth{Xygp}%
  \raisebox{-\mydepth}{\includegraphics[height=\myheight]{#1}}%
}

\def\BibTeX{{\rm B\kern-.05em{\sc i\kern-.025em b}\kern-.08em
    T\kern-.1667em\lower.7ex\hbox{E}\kern-.125emX}}
\begin{document}

\title{Grounding Hindsight Instructions in Multi-Goal Reinforcement Learning for Robotics \\
{\footnotesize Published at the 2022 IEEE International Conference on Development and Learning (ICDL)}
\thanks{The authors gratefully acknowledge funding by the German Research Foundation DFG through the IDEAS (402776968) and LeCAREbot (433323019) projects.}
}

\author{\IEEEauthorblockN{1\textsuperscript{st} Frank Röder,}
\IEEEauthorblockA{\textit{Institute for Data Science Foundations} \\
\textit{Hamburg University of Technology}\\
Hamburg, Germany \\
frank.roeder@tuhh.de}
\and
\IEEEauthorblockN{2\textsuperscript{rd} Manfred Eppe}
\IEEEauthorblockA{\textit{Institute for Data Science Foundations} \\
\textit{Hamburg University of Technology}\\
Hamburg, Germany \\
manfred.eppe@tuhh.de}
\and
\IEEEauthorblockN{3\textsuperscript{nd} Stefan Wermter}
\IEEEauthorblockA{\textit{Knowledge Technology} \\
\textit{University of Hamburg}\\
Hamburg, Germany \\
stefan.wermter@uni-hamburg.de}
}

\maketitle

\begin{abstract}
	This paper focuses on robotic reinforcement learning with sparse rewards for natural language goal representations.
	An open problem is the sample-inefficiency that stems from the compositionality of natural language, and from the grounding of language in sensory data and actions.
	We address these issues with three contributions. We first present a mechanism for hindsight instruction replay utilizing expert feedback. Second, we propose a seq2seq model to generate linguistic hindsight  instructions. Finally, we present a novel class of language-focused learning tasks.
	We show that hindsight instructions improve the learning performance, as expected. In addition, we also provide an unexpected result: We show that the learning performance of our agent can be improved by one third if, in a sense, the agent learns to \emph{talk to itself} in a self-supervised manner. We achieve this by learning to generate linguistic instructions that would have been appropriate as a natural language goal for an originally unintended behavior. Our results indicate that the performance gain increases with the task-complexity.

\end{abstract}

\begin{IEEEkeywords}
  reinforcement learning, language grounding, instruction following, hindsight instruction, human-robot interaction
\end{IEEEkeywords}

\section{Introduction}%
\label{sec:introduction}

Enabling robots to ground language in action has been ongoing research for decades \cite{Bisk_ExperienceGrounds_2020,Tellex_RobotsThat_2020,Lynch_LanguageConditioned_2021,Akakzia_DECSTR_2021}.
Language is often the most important communication channel for human-human and human-robot interaction ({HRI}).
More recently, researchers made progress in robotics, utilizing language to learn diverse behaviors and understand instructions within simulations \cite{Lynch_LanguageConditioned_2021} and the real world \cite{Shridhar_CLIPortWhat_2021}.
However, these learning approaches are still not on par with the zero-shot learning capabilities of human infants. Machine learning methods require huge amounts of offline data, large pre-trained models and many human-based language annotations.
In this article, we ask how we can improve the sample-efficiency and alleviate the need for language annotated robotic datasets.

\begin{figure}[!ht]
  \centering
  \includegraphics[width=0.5\linewidth]{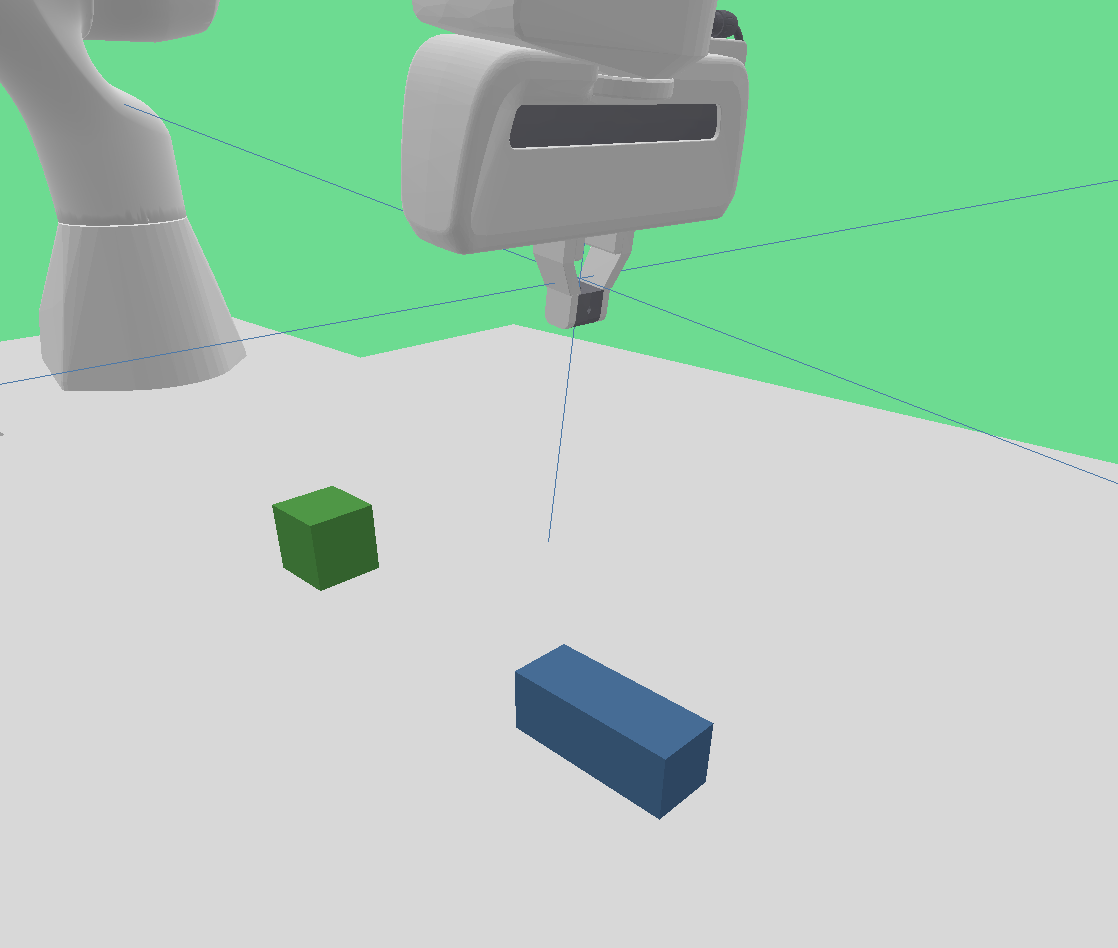}
  \caption{A scene from our simulated \envname{} environment with the Franka Emika Panda robot and two objects. An instruction like \say{reach the green cube} informs about the goal.}
  \label{fig:scene}
\end{figure}

In our recent review on embodied language learning \cite{Roder_EmbodiedCrossmodalSelf_2021}, cognitive principles emphasize the importance of linguistic feedback that toddlers experience while they explore in an intrinsically motivated fashion or imitate their social partners and caretakers.
Here, we want to transfer these findings to {HRI}, where natural language allows describing goals and conditions.
An important feature of natural language is its compositionality: it enables the capability to change the context of an utterance in the presence of a wrong outcome.
For example, consider the scene in \autoref{fig:scene}, where the robot's task could be to \say{touch the green object} (\inlinegraphics{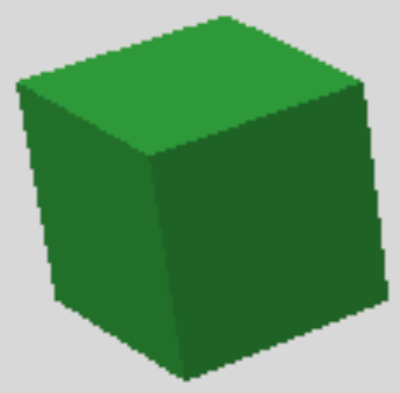}).
Since the object is no further defined, the color-difference determines the correct understanding, and all other properties can be ignored.
Similarly, an example with the task to \say{reach the cuboid} (\inlinegraphics{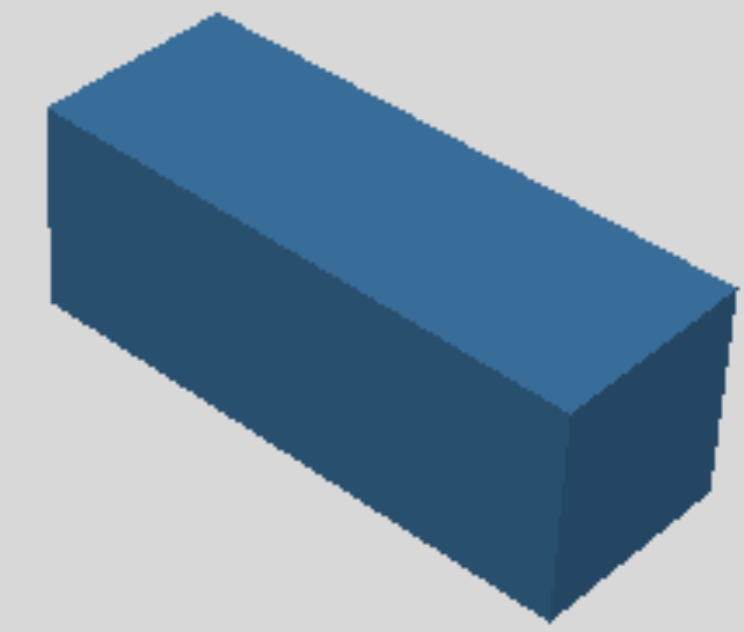}) can be considered respectively.
This, combined with an external teacher to signal whether instructions have been followed correctly, enables a possibility to ground language about actions, colors and shapes in sensorimotor experience.
However, a problem is that the natural language instructions are often misunderstood, leading to erroneous robotic behavior.
The problem that we address in this article, is to learn from misunderstood instructions.

A promising solution to address potential failures in robotic learning is hindsight learning\cite{Andrychowicz_HER_2017,Eppe_IntelligentProblemsolving_2022}.
Hindsight methods enable learning from undesired outcomes by treating them as desired ones.
Technically, they relabel a present behavior with a goal or instruction that was actually satisfied.
A well-known approach is \textit{Hindsight Experience Replay} (HER) \cite{Andrychowicz_HER_2017} which solved many multi-goal reinforcement learning ({RL}) tasks with sparse rewards that were entirely unsolvable or where contemporary methods achieved only poor performing solutions.
However, {HER} was initially designed for goals-as-states and is not trivial to adapt for linguistic instructions.
This shortcoming has been addressed by the implementation of {HIGhER} \cite{Cideron_HIGhERImproving_2020}, where the authors exploit the reward signal to collect positive behavior-instruction samples.
They use these pairs to train a discriminative model, which predicts instructions that appropriately describe a goal state in a discrete grid world environment.

However, there exists no research beyond the grid world, towards continuous robotic environments with realistic dynamics, where an approach akin to {HIGhER} \cite{Cideron_HIGhERImproving_2020} is used.
This is more challenging because in robotics, the mapping between desired goal states and appropriate instructions to achieve them is $\infty$-to-one. For example, consider the complex context-specific continuous-world behavior related to instructions involving \textit{grasping} and \textit{pushing}.
In this article, we address this gap with the following three contributions:

\begin{itemize}
	\item \textit{Hindsight Expert Instruction Replay} (\heir{}) is a replay mechanism to relabel instructions that led to unintended behavior with hindsight expert instructions that are appropriate for the unintended behavior.
	For example, consider that an agent has been instructed to \say{reach the green cube}, but it reaches the blue one. Then \heir{} is a method that uses a hard-coded environment-specific language generator to relabel a training episode with a retrospectively correct instruction, which in this case would be \say{reach the blue cube}.
	\item \textit{Hindsight Instruction Prediction from State Sequences} (\hipss{}) replaces the hard-coded environment-specific language generator with a seq2seq learning mechanism to generate the hindsight expert instructions. \hipss{} is self-supervised, in the sense that it automatically generates training data based on the reward signal and grammatical constraints as background knowledge.
	\item \textbf{LAN}guage \textbf{RO}botics (\envname{}), a novel class of environments for language-conditioned {RL} with synthetic instructions and hindsight expert features.
\end{itemize}

We hypothesize that external hindsight instructions generated by \heir{} allow an agent to exploit the compositionality of language by accelerating and stabilizing the training procedure.
Inspired by {HIGhER} \cite{Cideron_HIGhERImproving_2020}, we also hypothesize that the training procedure can be accelerated if the hindsight instructions are self-generated with our \hipss{} method.
To address these hypotheses, we extend the widely used and standardized interface of OpenAI Gym \cite{Brockman_OpenAIGym_2016} with linguistic goal representations and hindsight instructions (\envname{}).

In the remainder of this article, we first highlight the current approaches to language-conditioned and hindsight learning in \autoref{sec:related_work}.
With \autoref{sec:background}, we provide the theoretical background in goal-conditioned problem-solving.
Then, \autoref{sec:methodology} presents our implementation details, followed by our experimental setup and results in \autoref{sec:experiments}.
We close the article with a discussion in \autoref{sec:discussion} and a short conclusion in \autoref{sec:conclusion}.

\section{Related Work}%
\label{sec:related_work}

\subsection{Language Grounding in Reinforcement Learning}%
\label{sub:language_grounding_in_reinforcement_learning}

In the review of Luketina et al. \cite{Luketina_SurveyRL_2019}, the authors highlight the applicability of language in {RL}, not only as another type of goal definition but also as a source of information for problem-solving.
To this date, most of the studies investigated skill learning with sparse rewards \cite{Andrychowicz_HER_2017}, while only some of them incorporate language as part of the observation \cite{Jiang_HAL_2019,Akakzia_DECSTR_2021,Hill_GroundedLanguage_2021}, enabling language grounding based on rewards.

One exceptional example using language to describe goals in {RL} is the \textit{BabyAI} platform \cite{Chevalier-Boisvert_BabyAI_2019}.
It provides integrated expert demonstrations, causal puzzles, human-in-the-loop capabilities, and a synthetic subset of the English language, called \textit{BabyLanguage}.
In environments like the \textit{DeepMind Lab} \cite{Hill_GroundedLanguage_2021}, the agent is situated in a simple simulated world and instructed to solve navigation tasks using a narrow action space.
Furthermore, there exists experiments with language grounding in video games like \textit{ViZDoom} \cite{Chaplot_Gated-attention_2018}.
However, video games often have a very limited control and sensor space.
Based on our previous surveys \cite{Roder_EmbodiedCrossmodalSelf_2021,Eppe_IntelligentProblemsolving_2022}, we assume that a physically realistic interaction space is crucial for advances in language grounding because it involves continuous actions to explore concepts such as \textit{light} or \textit{heavy} and complex motions like \textit{grasping} or \textit{pushing}.
Such concepts are vital for instruction-following in {HRI} \cite{Tellex_RobotsThat_2020,Lynch_LanguageConditioned_2021}.
However, most existing language-based robotic applications only learn from successful trials \cite{Jiang_HAL_2019,Hill_GroundedLanguage_2021}, while only some of them \textit{retry on failure} \cite{Lynch_LanguageConditioned_2021} and are capable of \textit{logically} combining multiple requests \cite{Akakzia_DECSTR_2021,Shridhar_CLIPortWhat_2021}.
To investigate language-conditioned {RL}, we consider established methods for goal-conditioned RL \cite{Schaul_UVFA_2015,Haarnoja_SoftActorCritic_2018,Andrychowicz_HER_2017}.

\section{Background}%
\label{sec:background}

In this section, we briefly discuss the methodology of RL, especially multi-goal RL and language-conditioned RL.
As a central part of this work, we introduce the formal notation of hindsight learning for goals and instructions.
\subsection{Reinforcement Learning}%
\label{sub:reinforcement_learning}
We use the well-known definition of the Markov decision process ({MDP}), found in the book by Sutton et al. \cite{Sutton_RL_Introduction_2018}.
In this work, the {MDP} is discrete in time; hence, we annotate the elements with a subscript to specify the current time step $_{t}$ and its successor $_{t+1}$.
Furthermore, the transition function of the simulation is deterministic.
The objective is to find a policy $\policy$ that selects actions, $a \sim \policy(s)$, to maximize the expected cumulative discounted reward $\E_{\policy} \left[ \sum_{t=0}^{\infty} \discount^{t} r_t \right]$.

\subsection{Multi-goal and language-conditioned RL}%
\label{sub:multi_goal_rl}
For the multi-goal setting, we augment the {MDP} by an additional set of goals $\gG$ \cite{Andrychowicz_HER_2017}, which are part of the state-space, therefore $\gG \subset \gS$.
This property allows to map any state $s \in \gS$ to a goal $g$, that we define as mapping function $m(s): \gS \rightarrow \gG$.
We modify the traditional reward function to compute the reward based on the episodic goal
\begin{equation}
  \label{eq:goal_conditioned_reward}
  r_{t} = R_g(s_{t+1}, g) =
  \begin{cases}
    0,  & \text{if} \; \|m(s_{t+1}) - g\|_{2} \leq \eps \\
    -1, & \text{otherwise}
  \end{cases}
\end{equation}
, where $\eps$ is a task-specific threshold.
The reward function $R_g(s_{t+1}, g)$ considers the achieved goal state $s_{t+1}$ and the desired episodic goal $g$ as input.
To apply this notion to linguistic goals $g_\ell$, we remove the mapping function $m(s)$ and replace the if-clause of \autoref{eq:goal_conditioned_reward} with a task-specific condition $C(s_{t+1},g_\ell)$ that is true if one of the many possible states of $\gS_{g} \subset \gS$ ($g_\ell$ describes) is achieved with a transition $(s_t, a_t, s_{t+1})$ leading to a state $s_{t+1} \in \gS_{g}$:
\begin{equation}
  \label{eq:language_conditioned_reward}
  r_{t} = R_{g}(s_{t+1}, g_\ell) =
  \begin{cases}
    0,  & \text{if} \; C(s_{t+1}, g_\ell) \\
    -1, & \text{otherwise}
  \end{cases}
\end{equation}.

In both cases, we assume to sample a state-goal pair $\rho(s_0, g)$ or state-instruction pair
$\rho(s_0, g_\ell)$ respectively at the beginning of each episode, while $\rho$ defines the distribution of start configurations.
From the work of \textit{Universal Value Function Approximators} \cite{Schaul_UVFA_2015}, we know that all the theoretical formulations around the Bellman equation still hold, and we can learn universal policies that can solve linguistic goals if not any type of goal representation \cite{Nair_VisualReinforcement_2018}.

\subsection{Hindsight RL}%
\label{sub:hindsight_rl}
Hindsight experience replay \cite{Andrychowicz_HER_2017} usually assumes access to the function of \autoref{eq:goal_conditioned_reward} to relabel trajectories by transforming visited states into goals \textit{a posteriori}.
Formally, the agent experiences a sequence of transitions $\left\{ (s_0,a_0,r_1,s_1), \ldots (s_{T-1},a_{T-1},r_{T},s_{T}) \right\}$ up to the episode limit of $T$.
For the terminal state $s_T$, given a goal $g$, one can evaluate the reward function from \autoref{eq:goal_conditioned_reward}.
In case of a penalty $r_T = -1$, $g$ could be replaced with $g' = m(s_T)$, the corresponding goal that the final state maps to.
This cannot be directly used for language-conditioned {RL}, as it assumes every visited state $s_t$ to be a potential goal.
Unlike previous articles \cite{Jiang_HAL_2019,Cideron_HIGhERImproving_2020,Akakzia_DECSTR_2021}, we address this issue by using the condition of \autoref{eq:language_conditioned_reward} to detect interactions with any object, therefore linguistically relevant states.

\section{Methodology}%
\label{sec:methodology}

For our agent implementation, we use the \textit{Soft Actor Critic} ({SAC}) \cite{Haarnoja_SoftActorCritic_2018}, a state-of-the-art off-policy algorithm commonly used in RL research for continuous perception and control.
The implementations \textit{HAL} \cite{Jiang_HAL_2019} and \textit{decstr} \cite{Akakzia_DECSTR_2021} are examples using {SAC} for instruction-following.

Hindsight learning alleviates the imbalance of successful and unsuccessful trials within the agent's replay memory, by relabeling the failures with appropriate instructions and sparse rewards in hindsight.
This augmentation technique based on contrived transitions not only improves the sample-efficiency, but also the language learning performance \cite{Jiang_HAL_2019,Akakzia_DECSTR_2021}.
Similar to language development of toddlers \cite{Roder_EmbodiedCrossmodalSelf_2021}, we use two methods to obtain the hindsight instructions.

\subsection{Expert Feedback}%
\label{sub:expert_feedback}
Prior works \cite{Chevalier-Boisvert_BabyAI_2019,Colas_IMAGINE_2020,Akakzia_DECSTR_2021} have shown, that a social partner \cite{Akakzia_DECSTR_2021} or task expert \cite{Nguyen_InteractiveLearning_2021} amends the learning of an artificial agent.
They can provide linguistic feedback, as they have the knowledge to label the observed behavior with a suitable instruction after the fact.
Formally, an expert returns an instruction for a sequence of states $(s_0,s_1, \ldots, s_{t}) \rightarrow g_{l}$, the hindsight linguistic goal.
In this paper, we use a mechanism derived from {HER} \cite{Andrychowicz_HER_2017}, called \heir{} with 3 replay strategies:

\begin{itemize}
  \item \textit{episode}: Given an episode with a hindsight expert signal at time $t$, we randomly select $k$ transitions within $i$, while $0 \leq i \leq t \; \forall i$, and relabel them with the hindsight instruction and a sparse reward of $0.0$ if $i = t$ or a reduced penalty of $-0.9$ otherwise. This should reinforce the agent to rate those transitions as valuable in satisfying the hindsight goal instruction.
  \item \textit{future}: For a given episode with a hindsight expert transition at time $t$, we randomly select $k$ transitions from $i$, while $t \leq i < T \; \forall i$, and relabel them with hindsight instructions and a sparse reward, if $i=t$ or a reduced penalty otherwise. This should incentivize the agent to consider the transitions after the hindsight signal of high quality, as they are similar to the desired goal state or preferable aftereffect of a successful behavior.
  \item \textit{final}: For an episode and the hindsight signal at time $t$, we consider the last transition at time step $t$ and the immediate $k$ predecessors back to $t-k$ as replay candidates.
    In both cases, we replace the instruction with our hindsight instruction and substitute the reward with a sparse reward or a lowered penalty, respectively.
\end{itemize}

\subsection{Discriminative Approach}%
\begin{figure}[ht]
  \centering
  \includegraphics[width=1.0\linewidth]{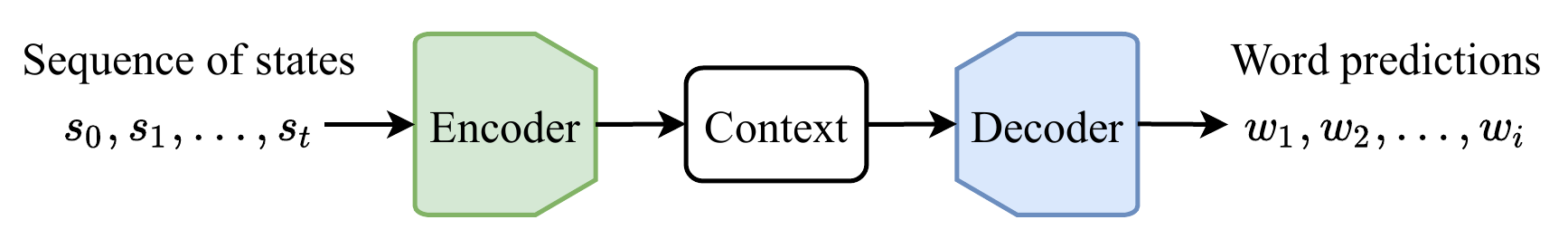}
  \caption{HIPSS model architecture. Both the encoder and decoder are multi-layered gated recurrent units \cite{Cho_GRU_2014}.
}
  \label{fig:hipss}
\end{figure}

\label{sub:discriminative_approach}
The agent uses a discriminative model to relabel unsuccessful sequences of states, by generating instructions in hindsight.
A similar idea had been implemented with {HIGhER} \cite{Cideron_HIGhERImproving_2020}, a method that learns this function from data labeled by the sparse reward.
Positive samples, where the trajectory coincides with proposed instruction, are used for training the model next to the policy.
However, their implementation is not publicly available, and the architecture described in their paper is limited to the 2D \textit{MiniGrid} environment \textit{BabyAI} \cite{Chevalier-Boisvert_BabyAI_2019}.
Another limitation is their focus on single transitions rather than the full trajectory.
We address this shortcoming with our seq2seq  model \cite{Sutskever_SequenceSequence_2014} \hipss{}, that predicts instructions based on sequences of states while also being available to the public.\footnote{Available at \sourcealgo{}}
\autoref{fig:hipss} depicts the architecture of \hipss{} and \autoref{algo:higher} describes the learning procedure in pseudocode.

The objective of \hipss{} is not enforcing to learn the grammar as we implicitly convey it with our word-based loss, which is sensitive to their particular positions in the instruction.
Formally, we train the model to estimate the conditional probability of words, given the trajectory of states seen in \autoref{eq:hipss}.
Here, $h$ represents the context (output of the encoder in \autoref{fig:hipss}) which the decoder uses to autoregressively generate the instruction.
\begin{equation}
  \label{eq:hipss}
  p( w_{1}, \ldots w_{N} | s_{0}, \ldots, s_{t} ) = \prod_{i=1}^{N} p(w_{i} | h, w_{1}, \ldots w_{i-1})
\end{equation}
Each word prediction is a distribution over all the words of our known task-specific vocabulary.

\begin{algorithm}[ht]
  \caption{\hipss{}}
  \label{algo:higher}
  \begin{tabular}[t]{l l @{\hspace{2.5em}} l l}%
    $\policy_{\theta}$    & policy & $M_{\phi}$ & \hipss{} model \\
    $\gB$ & replay buffer & $\gD$ & \hipss{} dataset\\
  \end{tabular}
  \begin{algorithmic}[1]
  \FOR{episode $i= 0 \ldots M$}
    \STATE Sample initial state $s_0$ and episodic goal $g_{\ell}^{i}$
    \FOR {time step $t= 0\ldots T$}
      \STATE Sample action $a_t \sim \policy_{\theta}(a_t | s_t, g_{\ell}^{i})$
      \STATE Take environment step $s_{t+1} \sim p(s_{t+1} | s_t, a_t)$
      \STATE Calculate reward $r_{t} \leftarrow R_{g}(s_{t+1}, g_{\ell}^{i})$
    \ENDFOR
    \IF{$r_{t} = 0$}
      \STATE Store successful trajectory $\gD \leftarrow \gD \cup{(s_{<t}, g_\ell^{i})}$
    \ELSIF{$r_{t} = -1$ and interplay with wrong object}
      \STATE Replace $g_{\ell}^{i}$ with predicted goal $\hat{g}_{\ell}^{i} \leftarrow M{\phi}(s_{<t})$
    \ENDIF
    \STATE Store transitions $\gB \leftarrow \gB \cup \left\{ (s_t, a_t, r_t, s_{t+1}, g_{\ell}^{i})^{T}_{t=0}\right\}$
    \STATE Update $\policy_{\theta}$ with mini-batch from $\gB$
    \STATE Update $M{\phi}$ with mini-batch from $\gD$
  \ENDFOR
  \end{algorithmic}
\end{algorithm}

\subsection{Summary}
\label{sub:method_summary}

\begin{figure}[ht]
  \centering
  \includegraphics[width=1.0\linewidth]{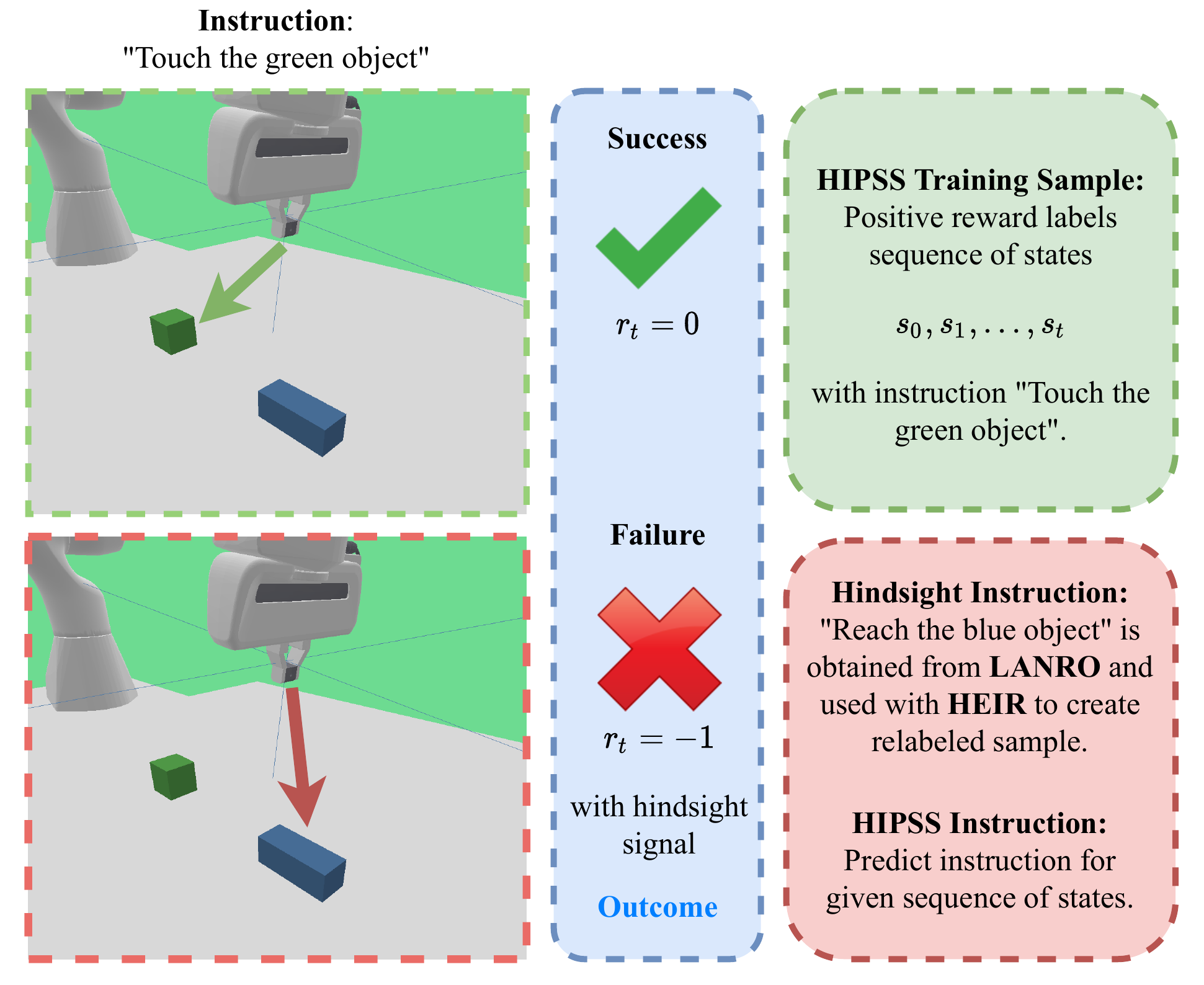}
  \caption{This figure presents our overall hindsight learning procedure. To the left, we have two scenarios given the instruction \say{Touch the green object}.
  On the top, the agent is following the instruction correctly (green arrow and framing) to obtain the sparse reward $r_t=0$. Subsequently, \hipss{} considers this as a positive training sample by adding the corresponding sequence of states and the instruction to its dataset. On the bottom with the red arrow and red framing, the agent receives a negative reward $r_t=-1$ next to a hindsight signal. While \heir{} uses the hindsight instruction \say{Reach the blue object} directly, \hipss{} takes the observed trajectory as input to generate an instruction on its own.}
  \label{fig:hindsight_learning}
\end{figure}

With \autoref{fig:hindsight_learning}, we depict the outcome-related mechanisms that are triggered for both methods \heir{} and \hipss{}, showing up the similarities but also their differences in terms of the origin of their hindsight instructions.
While \heir{} requires the synthetic social partner \cite{Akakzia_DECSTR_2021} to provide the instructions in hindsight, \hipss{} exploits the sparse reward from \autoref{eq:language_conditioned_reward} to collect positive training samples and to predict instructions by itself \cite{Cideron_HIGhERImproving_2020}.
In the following \autoref{sec:experiments}, we consider both approaches stand-alone.

\section{Experiments}
\label{sec:experiments}

\subsection{Environment}%
\label{sub:environment}

As of now, there is a lack of environments allowing low-level control in a robotic setup while providing hindsight instructions.
To address this, we present our language-focused learning environment \envname{} as a testbed for grounded language learning.\footnote{\envname{} got inspired by the implementation of \textit{panda-gym} \cite{Gallouedec_PandagymOpenSource_2021}}
In all tasks, we leverage a Franka Emika Panda robot with 7-DOF simulated by the PyBullet physics engine \cite{Coumans_PyBulletPython_2016}.
Following, we describe the essential components of our open-source learning platform.\footnote{Available at \sourceenv{}}

\subsubsection{States}%
\label{ssub:states}
The observed state consists of the agent's end effector state, joint positions, orientations, and velocities.
Furthermore, we add the object position, rotation, velocities, angular velocities, and object properties, such as the color or shaped, as concatenated one-hot encodings.

\subsubsection{Actions}%
\label{ssub:actions}
Actions are 4-dimensional vectors representing the relative change of the end effectors Cartesian position and the gripper state.

\subsubsection{Instructions}%
\label{ssub:instructions}
The episodic instruction is part of the state and encoded as a sequence of word indices based on the environment's vocabulary.

\subsubsection{Hindsight mechanism}%
\label{ssub:hindsight_mechanism}
Similar to prior work \cite{Chevalier-Boisvert_BabyAI_2019}, we generate hindsight instructions only for achieved states $s_{t+1}$ that prescribe an interaction with a wrong object which are distinct transitions within or at the end of an episode, utilizing \autoref{eq:language_conditioned_reward} for counterfactual evaluation.

\subsection{Task description}%
\label{sub:task_description}

For this paper, we consider the \textit{reach} task, where we instruct the robot to reach one out of two objects in a scene (cf. \autoref{fig:scene}).
The environment's collision detection evaluates the task-condition as true, when the robot slightly touches the correct object without causing a too large positional change to the goal and non-goal objects.
To study our implementations, we utilize $4$ modes to expand the space of instructions with respect to the number of color and shape words, while employing the verbs \textit{reach}, \textit{touch} and \textit{contact}:

\begin{enumerate}
  \item \mdefault{}: Instructions are generated based on a selection of 3 colors, \textit{red}, \textit{green}, and \textit{blue}. This option only contains objects of type \textit{box}. To expand the space of possible instructions, we use the 3 shape-specific words \textit{box}, \textit{block}, and \textit{square}, although the goal object could be identified by its color only.
  \item \mcolor{}: We expand the selection of colors by \textit{yellow}, \textit{purple}, \textit{orange}, \textit{magenta}, \textit{cyan}, and \textit{brown}, making the task more challenging while keeping the single shape.
  \item \mshape{}: We expand the \textit{Default} mode by a rectangle and cylinder, adding the shape-specific words, \textit{rectangle}, \textit{oblong} and \textit{brick}, and \textit{cylinder}, \textit{barrel}, and \textit{tophat} respectively.
  \item \mcolorshape{}: We combine the richness of colors from \mcolor{} with the diversity of objects from \mshape{}, to create our most demanding setup.
\end{enumerate}

In \autoref{tab:task_modes}, we provide an overview of the task modes with their total number of unique instructions the agent observes.

\begin{table}[!ht]
  \caption{The table lists the number of colors and shapes with the resulting number of total instructions per task mode.}
  \label{tab:task_modes}
	\centering
  \begin{tabular}{| l | c | c | c |}
  \hline
  Mode      & \#colors & \#shapes & \#instructions \\
  \hline
  \mdefault{}    & 3        & 1        & 9 \\
  \mcolor{}      & 9        & 1        & 27 \\
  \mshape{}      & 3        & 3        & 27 \\
  \mcolorshape{} & 9        & 3        & 81 \\
  \hline
  \end{tabular}
\end{table}

\subsection{Replay Strategy for Hindsight Instruction Replay}%
\label{sub:replay_strategy_for_hindsight_instruction_replay}

\begin{figure}[p]
  \centering
  \subfloat[\mdefault{}]{
    \includegraphics[width=0.4\textwidth]{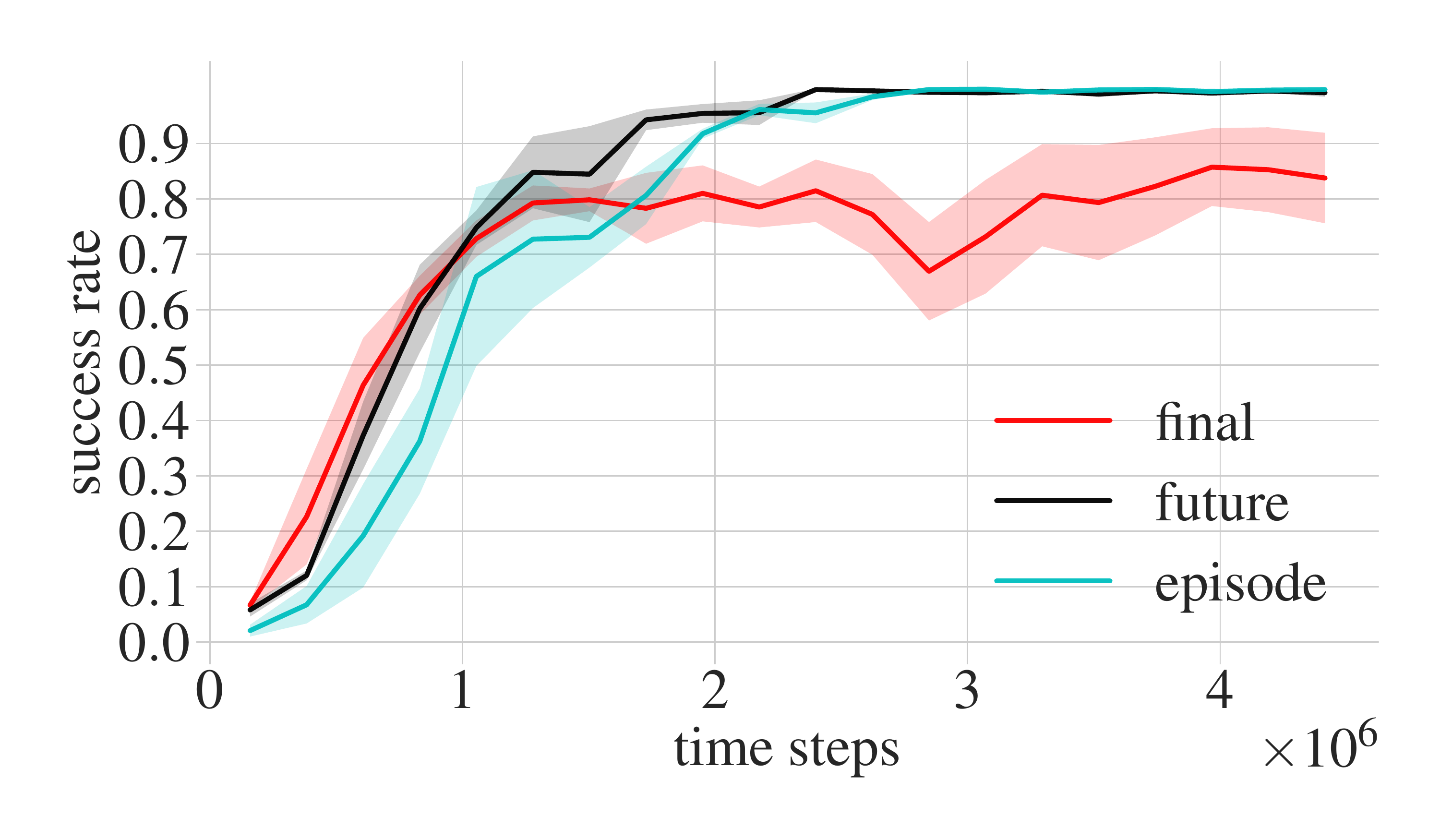}
    \label{fig:replay_strategies_default}
  }

  \subfloat[\mcolor{}]{
    \includegraphics[width=0.4\textwidth]{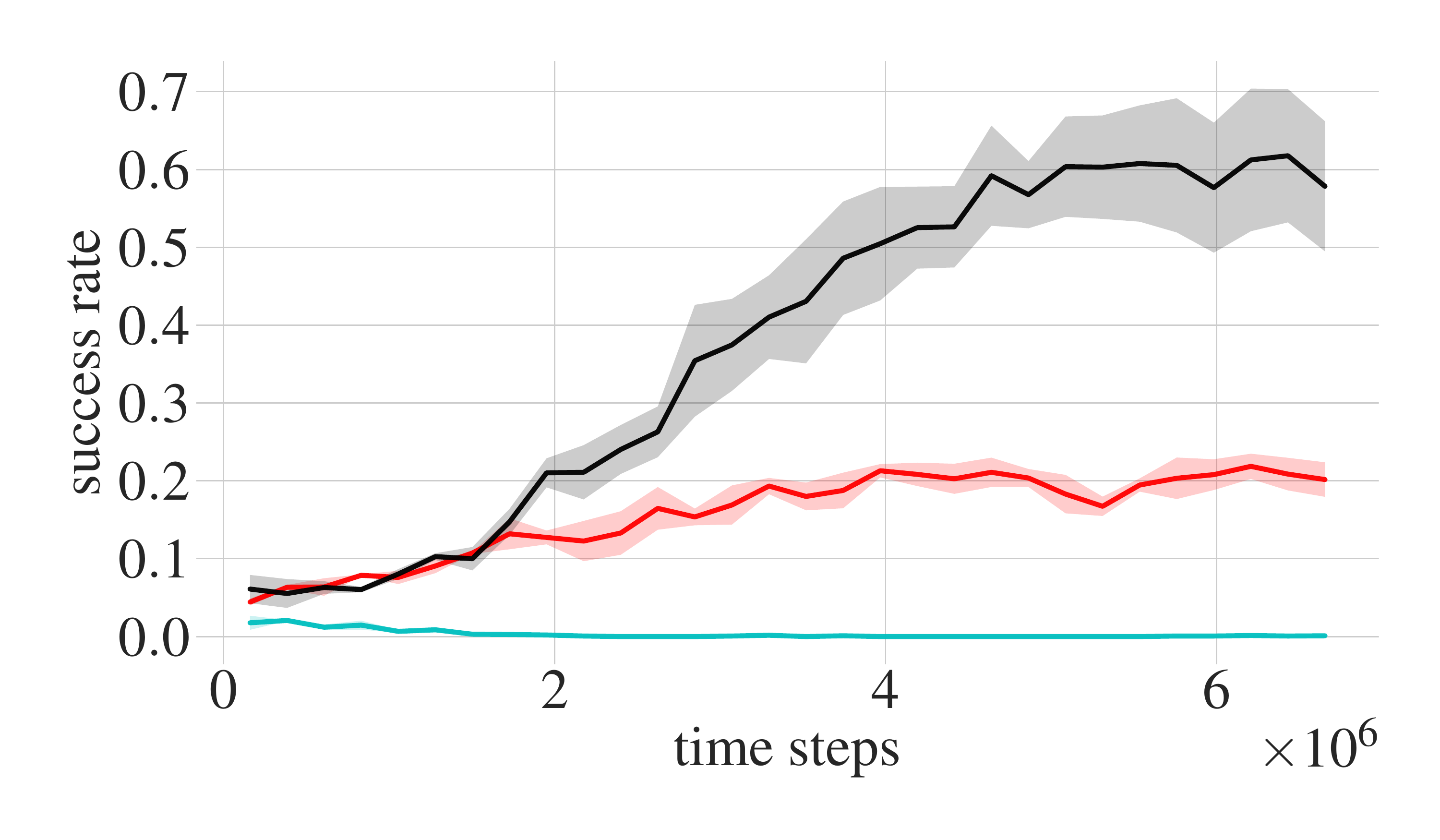}
    \label{fig:replay_strategies_color}
  }

  \subfloat[\mshape{}]{
    \includegraphics[width=0.4\textwidth]{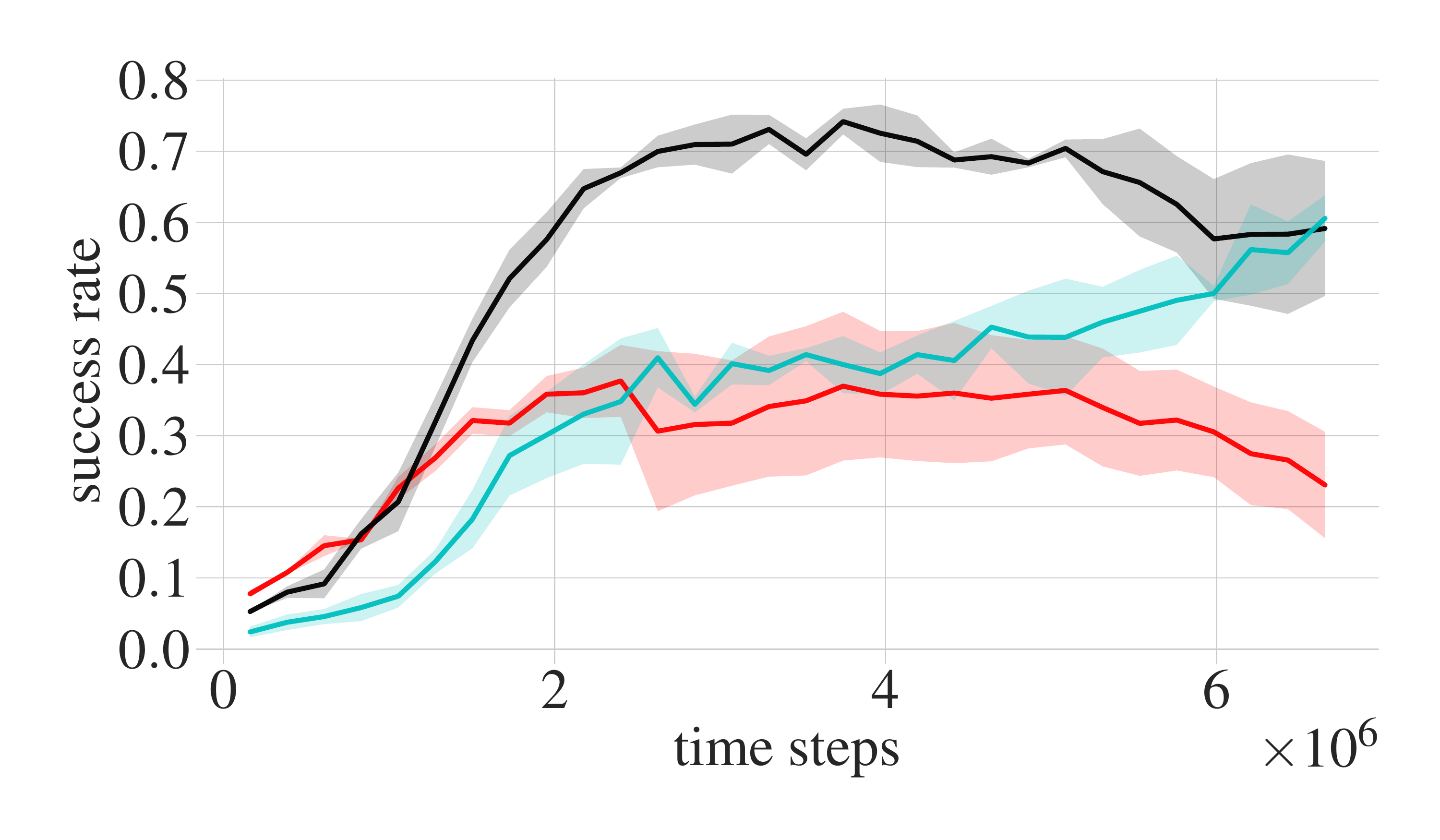}
    \label{fig:replay_strategies_shape}
  }

  \subfloat[\mcolorshape{}]{
    \includegraphics[width=0.4\textwidth]{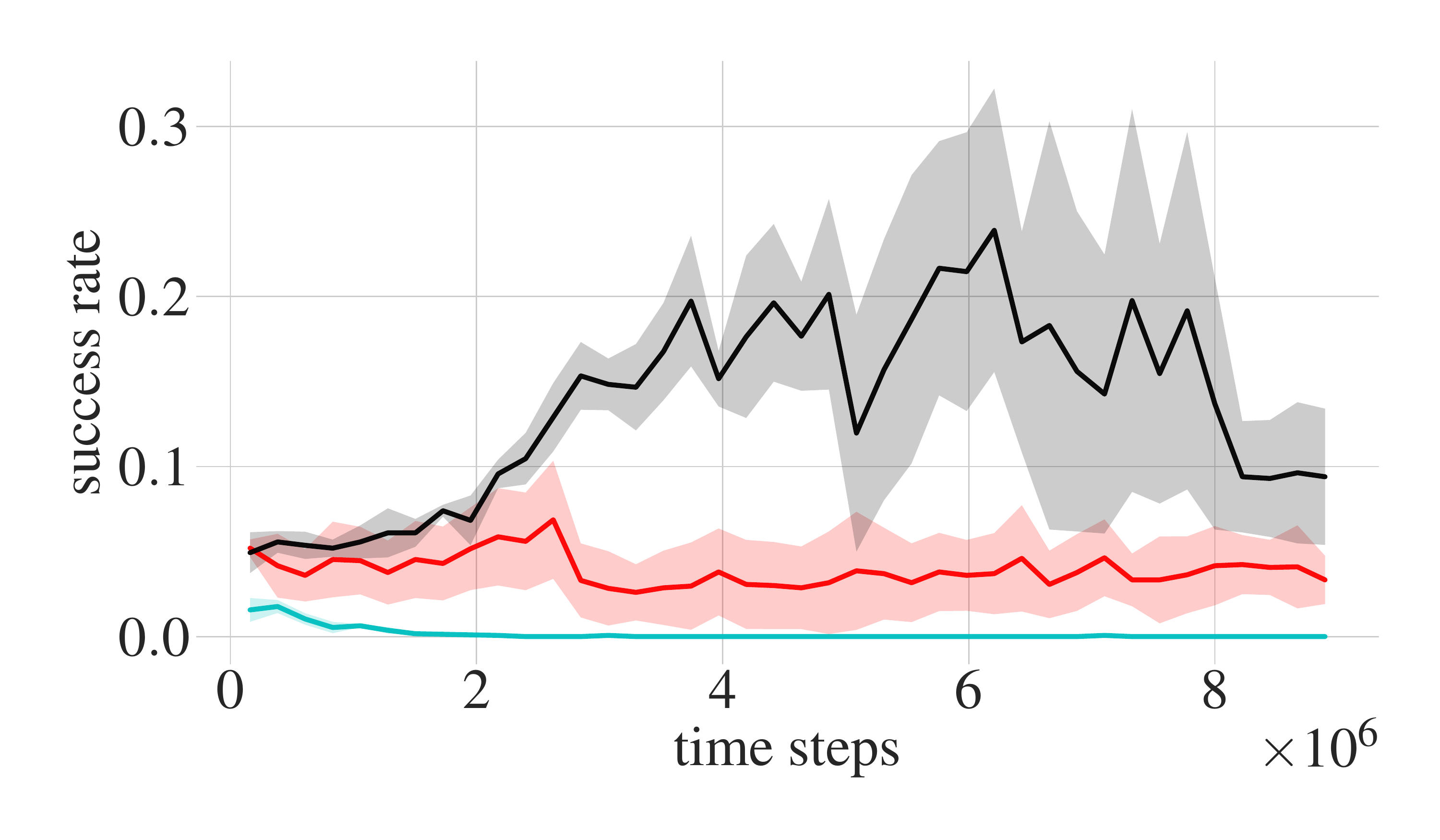}
    \label{fig:replay_strategies_color_shape}
  }
	\caption{The figure shows the success rate (y-axis) and the environment steps (x-axis) for the 4 task modes, contrasting the replay strategies \textit{future}, \textit{final}, and \textit{episode} (cf. \autoref{sub:expert_feedback}) with \heir{}. We consider the states close to an expert signal as valuable, and augment our replay buffer with those relabeled by a sparse reward (\texttt{0.0}) or a reduced penalty (\texttt{-0.9}) defined in \autoref{sub:expert_feedback}.}
  \label{fig:replay_strategies}
\end{figure}

Like the authors of {HER} \cite{Andrychowicz_HER_2017}, we empirically explored our 3 replay strategies by running experiments in all the 4 presented task modes of \autoref{sub:task_description}.
The results of \autoref{fig:replay_strategies} show the mean success rate of the presented strategies (\autoref{sub:expert_feedback}), with the shaded area as standard error of 3 trials for each configuration.

While one possibly expects the \textit{final} method to work the best, as it labels the transitions with the expert signal most of the time, the \textit{future} method performs at least twice as good because it also takes advantage of states reached afterward.
The \textit{episode} strategy performs the worst.
A reason for this could be the larger distance in terms of transitions until the hindsight signal appears.
All in all, our experiments favor the \textit{future} strategy, which is in accordance with the findings of the {HER} paper \cite{Andrychowicz_HER_2017}.
In the remainder of this article, we utilized the \textit{future} strategy to conduct experiments with both methods \heir{} and \hipss{}.

\subsection{Language Representations for Hindsight Learning}%
\label{sub:language_representations_for_hindsight_learning}

For our experiments, we use two methods to represent words as input to the actor and critic network, fed as sequence into a multi-layered gated recurrent unit \cite{Cho_GRU_2014}.
We tested one-hot encodings and trainable word embeddings as a lookup table of randomly initialized vectors for each word.
Unlike other methods \cite{Jiang_HAL_2019,Akakzia_DECSTR_2021}, we empirically confirm that learned embeddings outperform one-hot encodings (see \autoref{tab:word_embeddings}).
A possible explanation is that the embeddings capture the semantic similarities and relationship of task-relevant object properties in an end-to-end manner.
This is particularly useful for more complex linguistic instructions with several colors and shapes, which our hindsight procedure possibly exploits.

\begin{table}[!ht]
  \caption{This table showcases the mean episodic success of an agent trained with two different word representations, utilizing our two implementations, \heir{} and \hipss{}. The used representations are one-hot encodings and learned embeddings.}
  \label{tab:word_embeddings}
	\centering
  \begin{tabular}{| l | c | c | c | c|}
  \hline
  Mode & \heir{} & \hipss{} & \heir{} & \hipss{} \\
  \hline
                 & \multicolumn{2}{|c|}{one-hot} & \multicolumn{2}{|c|}{learned embeddings} \\
  \mdefault{}    & $48\pm 0.16$ & $61\pm 16$ & $82\pm 0.13$ & $\textbf{84}\pm 0.13$ \\
  \mcolor{}      & $30\pm 0.11$ & $16\pm 8$  & $\textbf{41}\pm 0.1$ & $37\pm 0.12$ \\
  \mshape{}      & $40\pm 0.18$ & $60\pm 15$ & $60\pm 0.1$ & $\textbf{66}\pm 0.13$ \\
  \mcolorshape{} & $12\pm 0.05$ & $14\pm 1$ & $22\pm 5$ & $\textbf{40}\pm 0.11$ \\
  \hline
  \end{tabular}
\end{table}

\begin{figure}[hp]
  \centering
  \subfloat[\mdefault{}]{
    \includegraphics[width=0.4\textwidth]{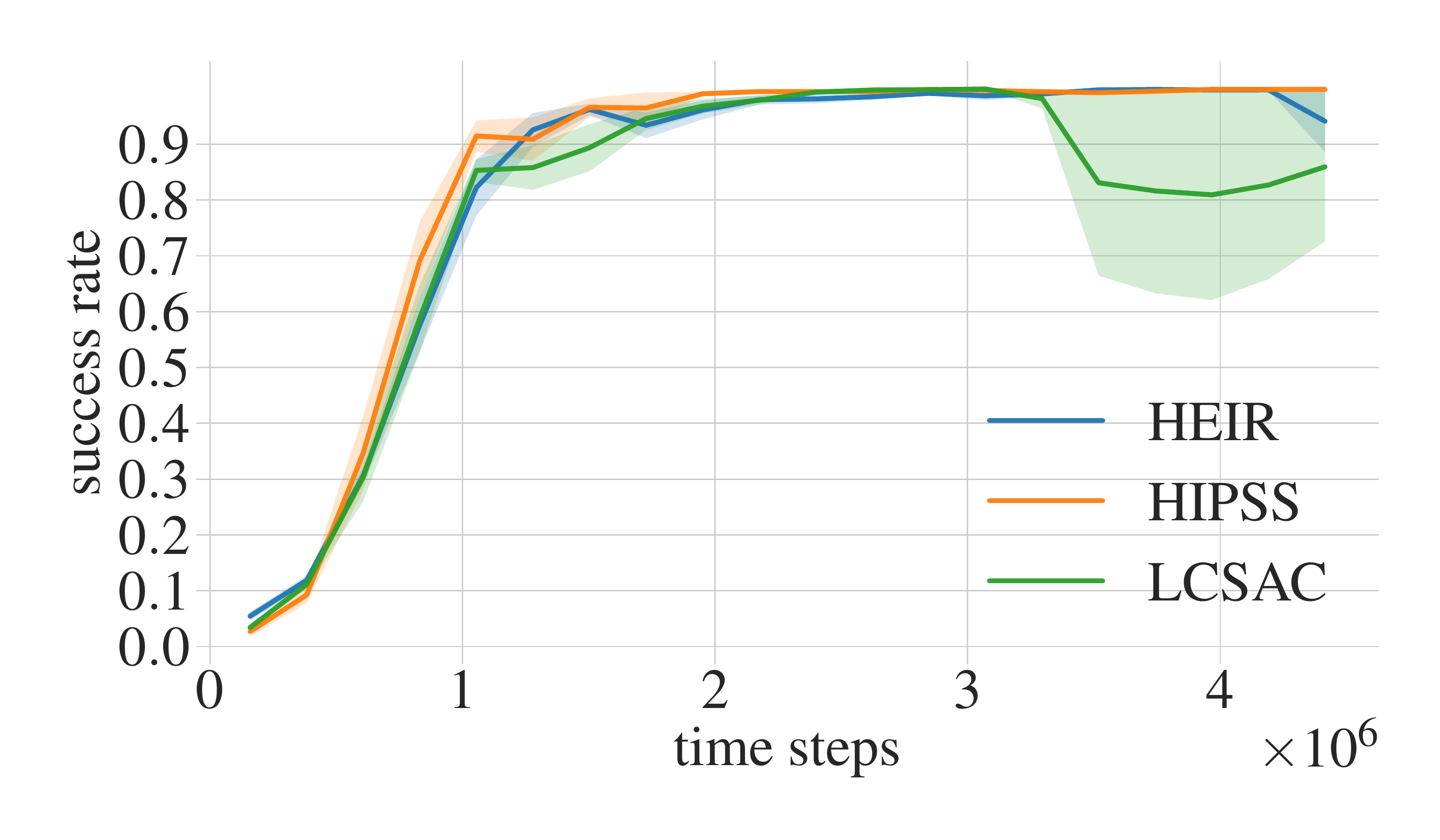}
    \label{fig:success_default}
  }

  \subfloat[\mcolor{}]{
    \includegraphics[width=0.4\textwidth]{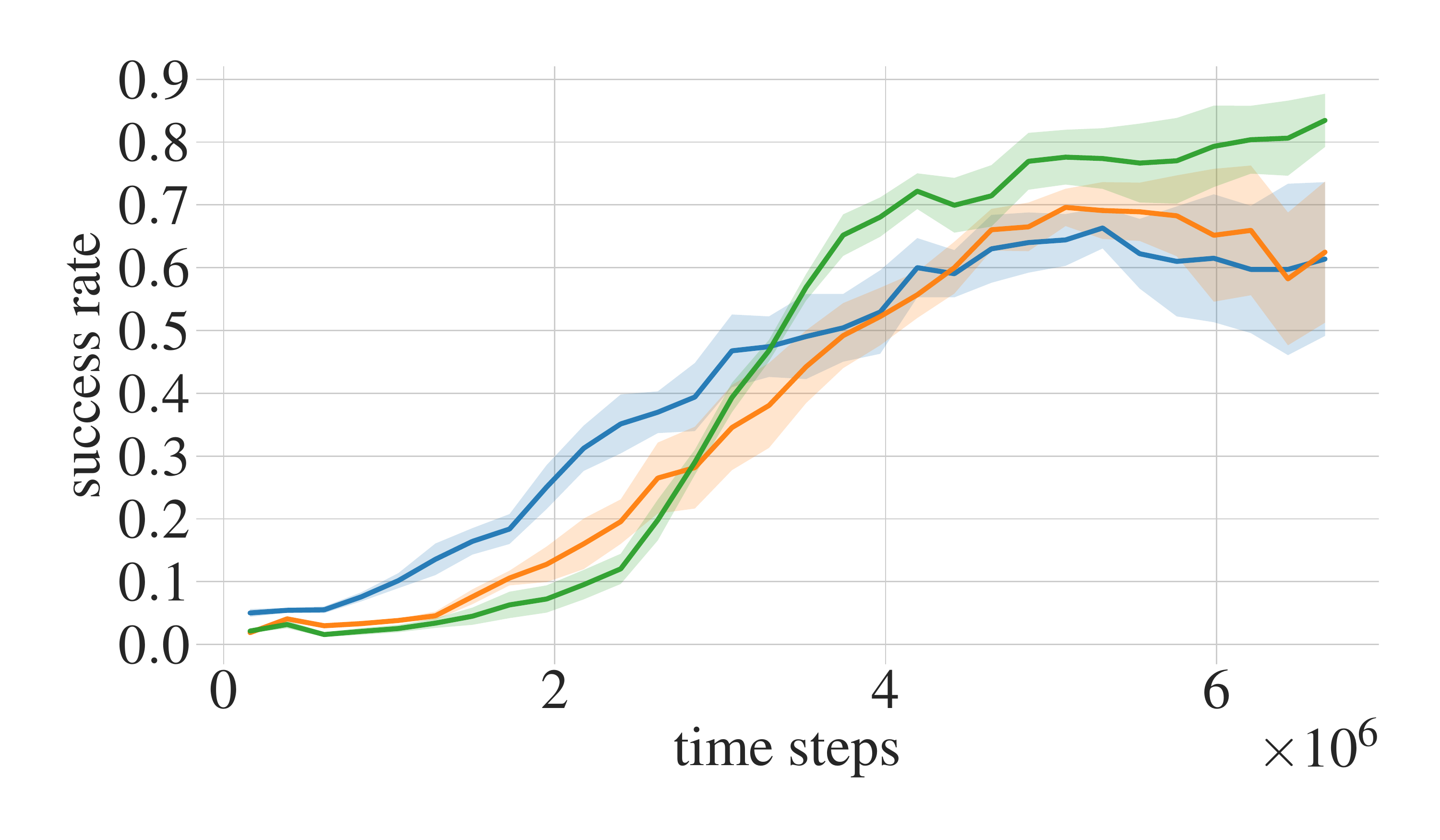}
    \label{fig:success_color}
  }

  \subfloat[\mshape{}]{
    \includegraphics[width=0.4\textwidth]{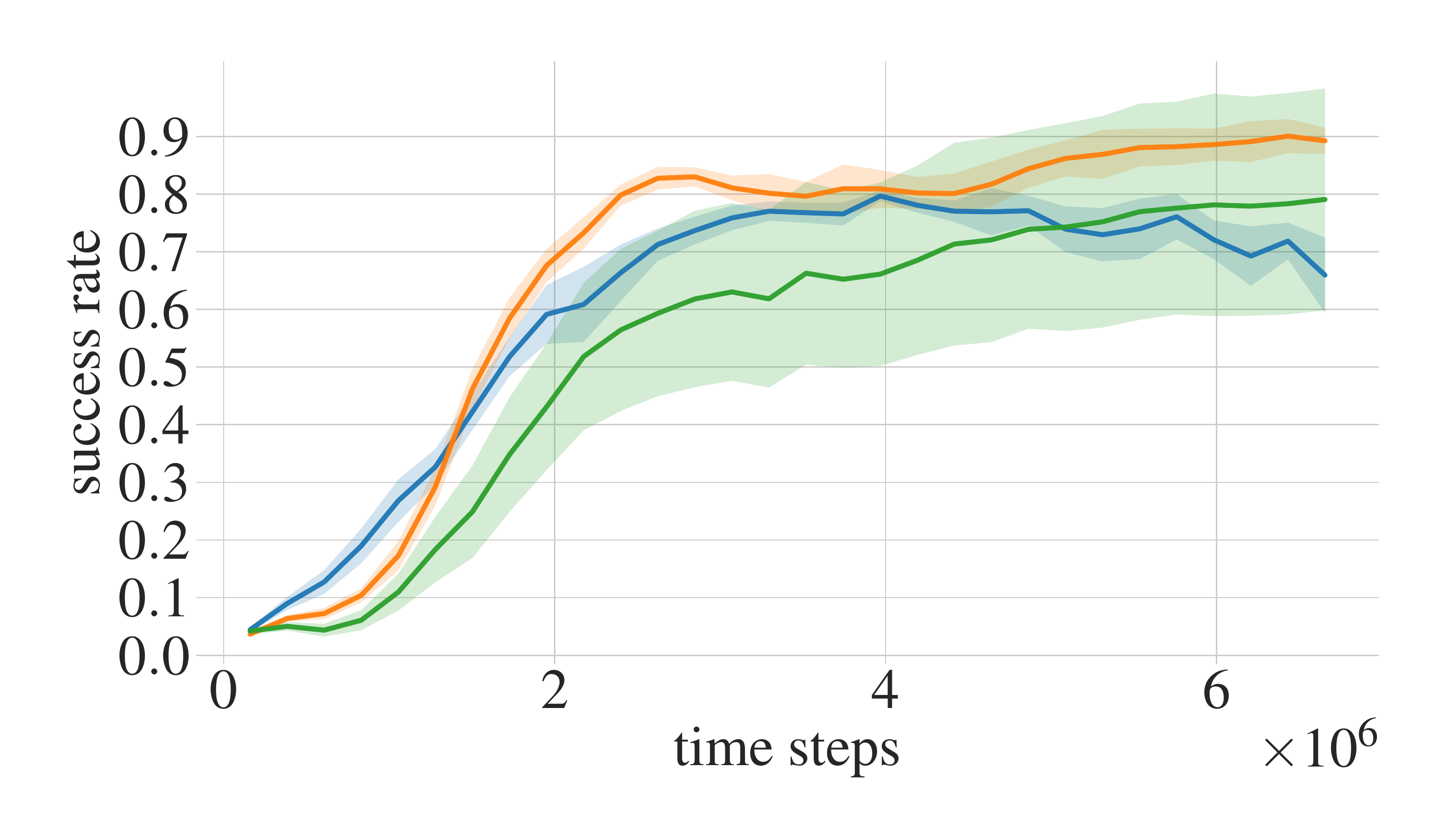}
    \label{fig:success_shape}
  }

  \subfloat[\mcolorshape{}]{
    \includegraphics[width=0.4\textwidth]{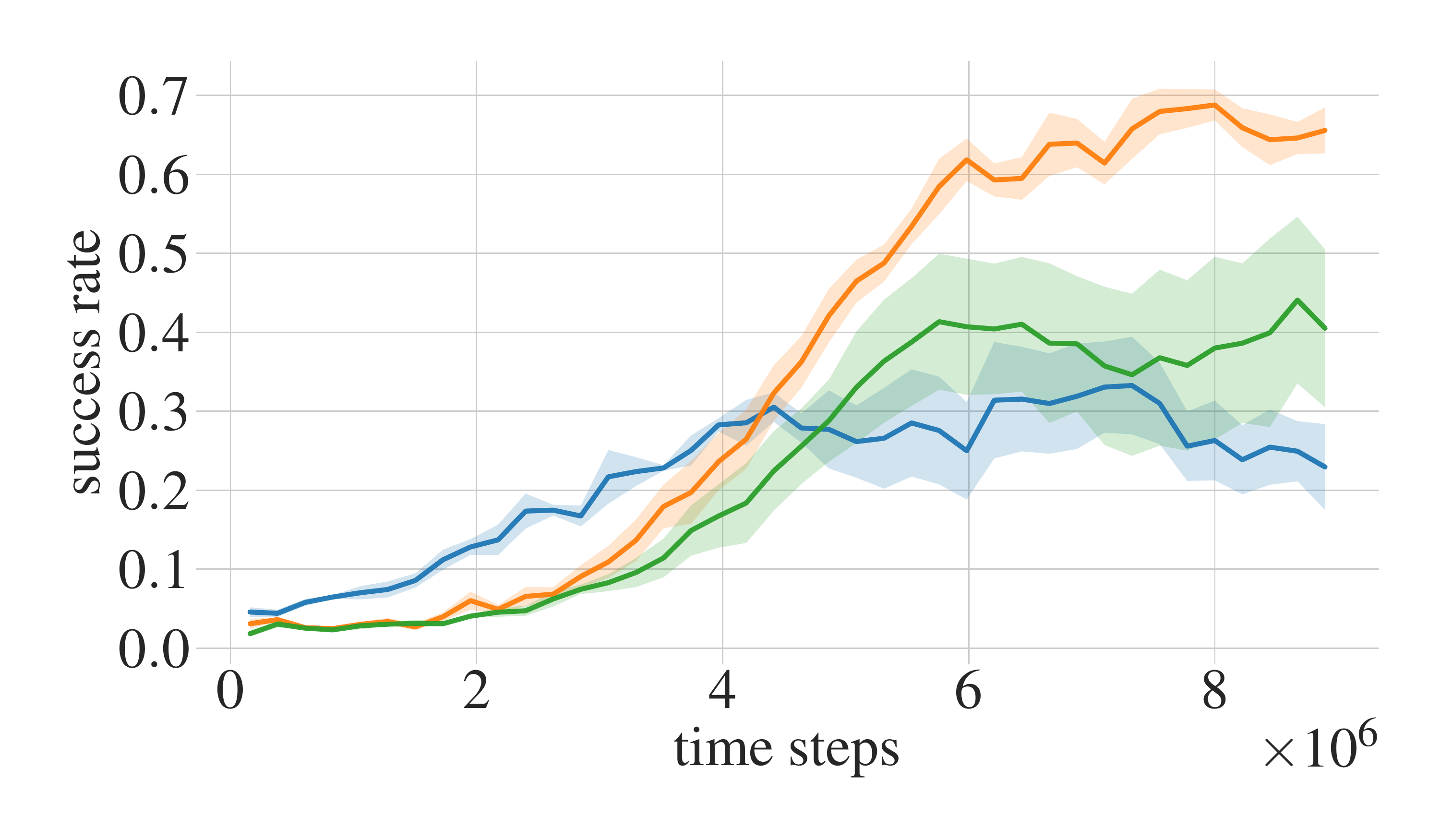}
    \label{fig:success_color_shape}
  }
  \caption{The figure shows the success rate (y-axis) and the environment time steps (x-axis) for the baseline \lcsac{}, \heir{}, and \hipss{}. Each configuration consists of 5 trials, with the shaded area being the standard error and the solid line the mean. We see performance gains for both hindsight methods in \autoref{fig:success_shape}, better convergence behavior for \hipss{} in \autoref{fig:success_default} and a better performance on average for the hardest setup in \autoref{fig:success_color_shape}.}
  \label{fig:results}
\end{figure}

\subsection{Results}%
\label{sub:results}

Our experiments compare the baseline language-conditioned {SAC} (\lcsac{}) to \heir{} and \hipss{}.
In \autoref{fig:results}, we use the \textit{reach} task with its 4 levels of difficulty to evaluate the language-conditioned learning capabilities with different amounts of task-relevant properties.
While the simple task mode \mdefault{} in \autoref{fig:success_default} does not convey major improvements except the improved stability in later training stages (around $3,500,000$ environment steps), \heir{} learns faster within the task \mcolor{} during the first $500,000$ steps, but then the learning performance of our approaches slows down compared to the baseline \lcsac{} (\autoref{fig:success_color}).\footnote{In the progress of training, the success rate rises and the policy makes fewer and fewer mistakes, which decreases the hindsight instruction generation.}
In \autoref{fig:success_shape} with the mode \mshape{}, \heir{} and \hipss{} show a better success rate than the baseline. They also indicate more stable learning properties because the deviation of the success rate is lower.
However, as illustrated in \autoref{fig:success_color_shape}, the most difficult task mode \mcolorshape{} shows other results.
The synthetic predictions generated with \heir{} seem not to improve the learning compared to \lcsac{}.
However, we observe major advantages when using \hipss{}.
It converges to a success rate of around 0.65, whereas the success rate of the baseline reaches 0.4 and \heir{} reaches 0.3.

We provide all the hyperparameters and single script to replicate the experimental results in our GitHub repository (cf. \autoref{sub:discriminative_approach}).

\section{Discussion}
\label{sec:discussion}
Our results show that self-predicted instructions with \hipss{} improve the training performance and overall success rate.
This is particularly surprising if we combine the data of \autoref{fig:results} with the considerations in \autoref{tab:task_modes}, which shows the linguistic complexity of the task modes: it indicates that our \hipss{} method provides the largest performance gain for the most difficult task modes.

\begin{figure}[hp]
  \centering
  \subfloat[\mdefault{}]{
    \includegraphics[width=0.4\textwidth]{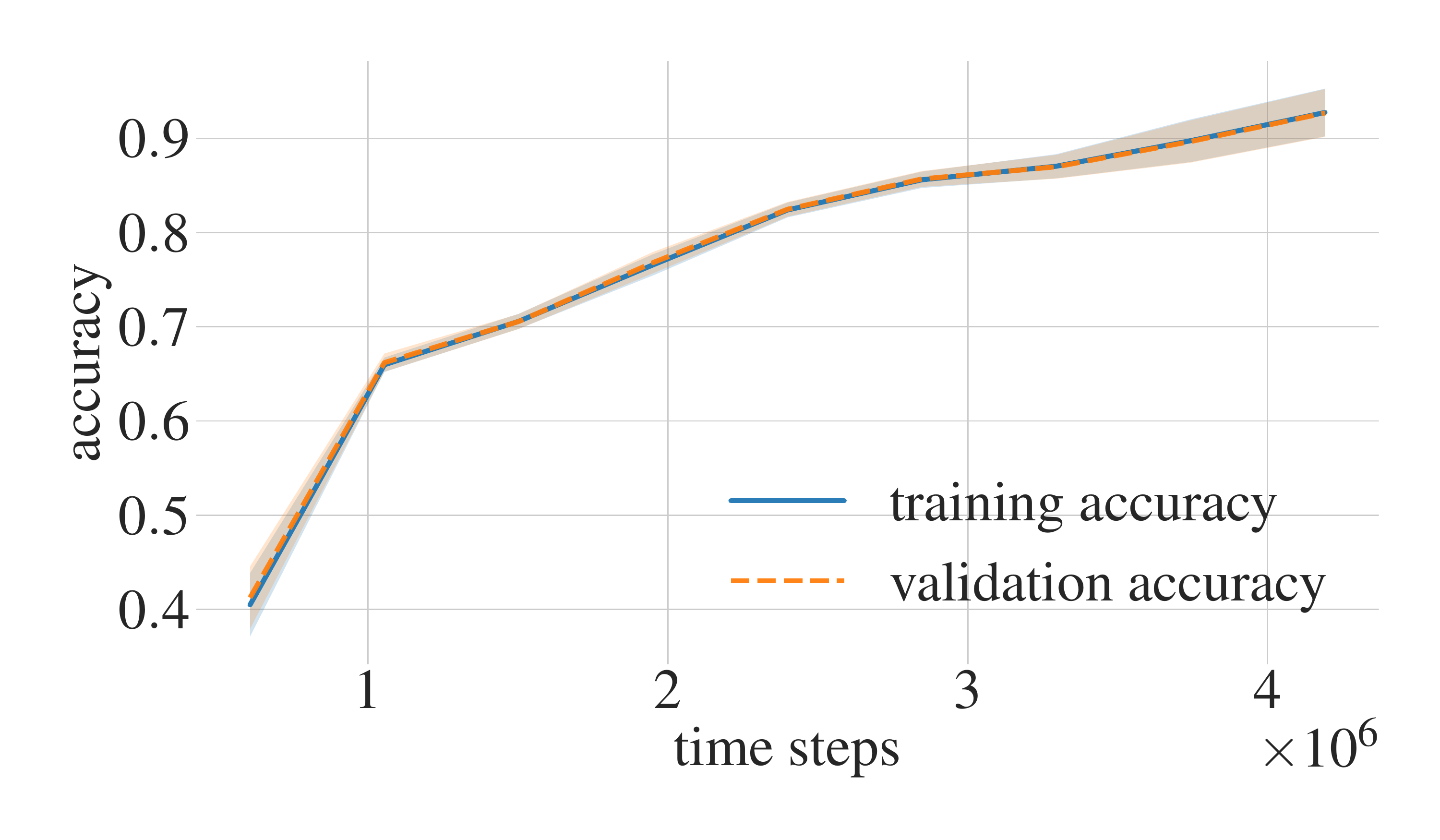}
    \label{fig:accuracy_default}
  }

  \subfloat[\mcolor{}]{
    \includegraphics[width=0.4\textwidth]{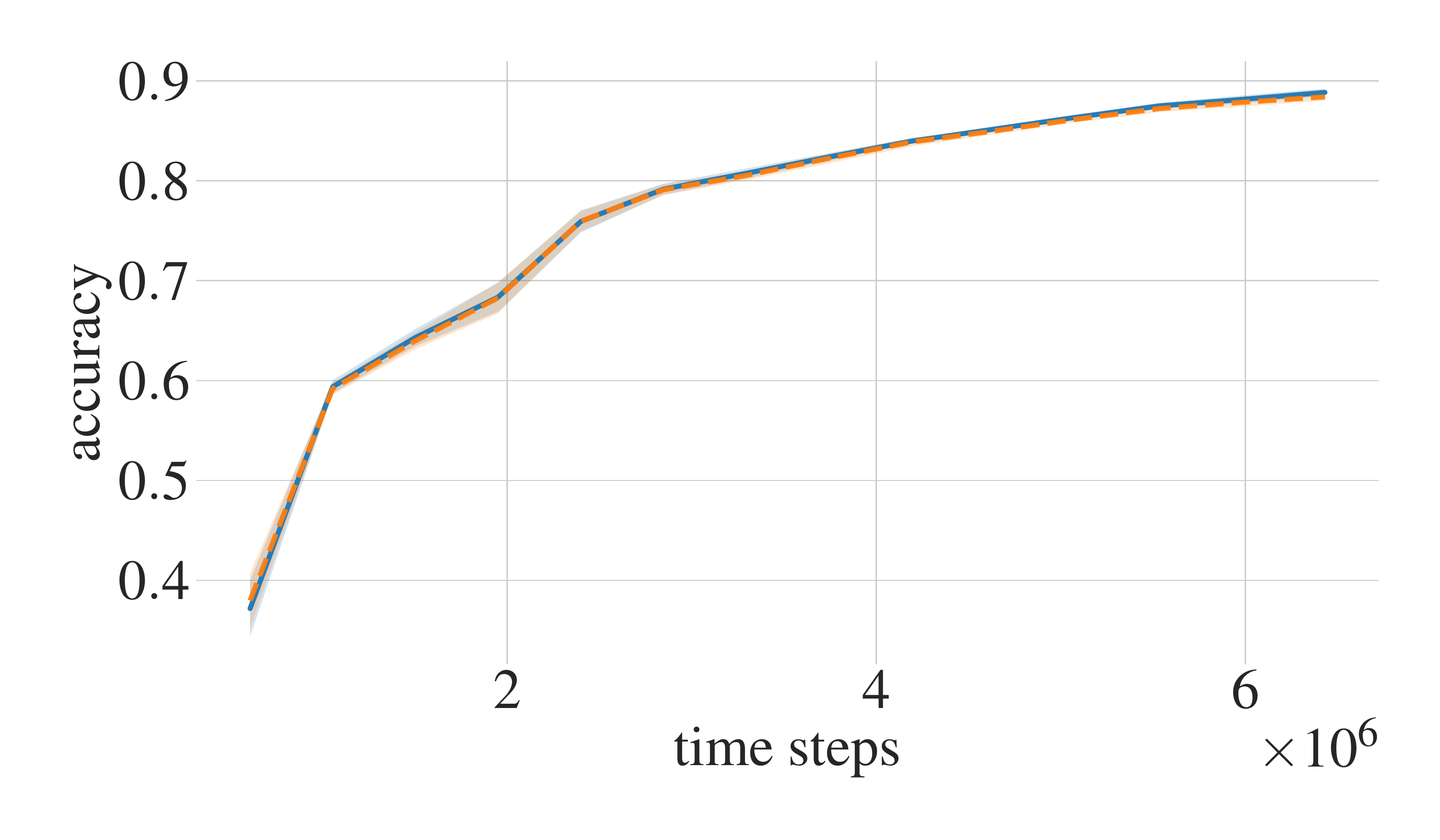}
    \label{fig:accuracy_color}
  }

  \subfloat[\mshape{}]{
    \includegraphics[width=0.4\textwidth]{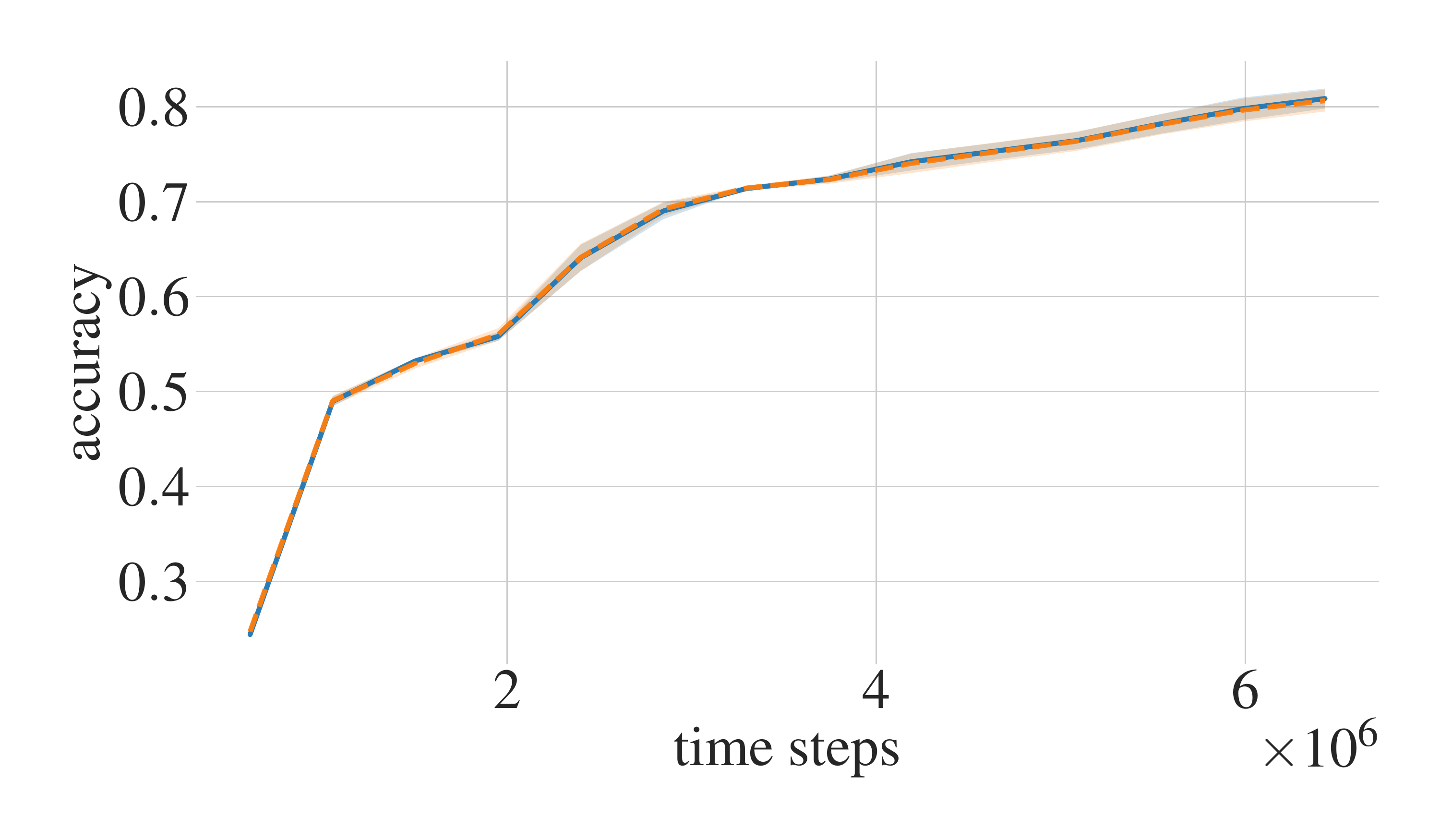}
    \label{fig:accuracy_shape}
  }

  \subfloat[\mcolorshape{}]{
    \includegraphics[width=0.4\textwidth]{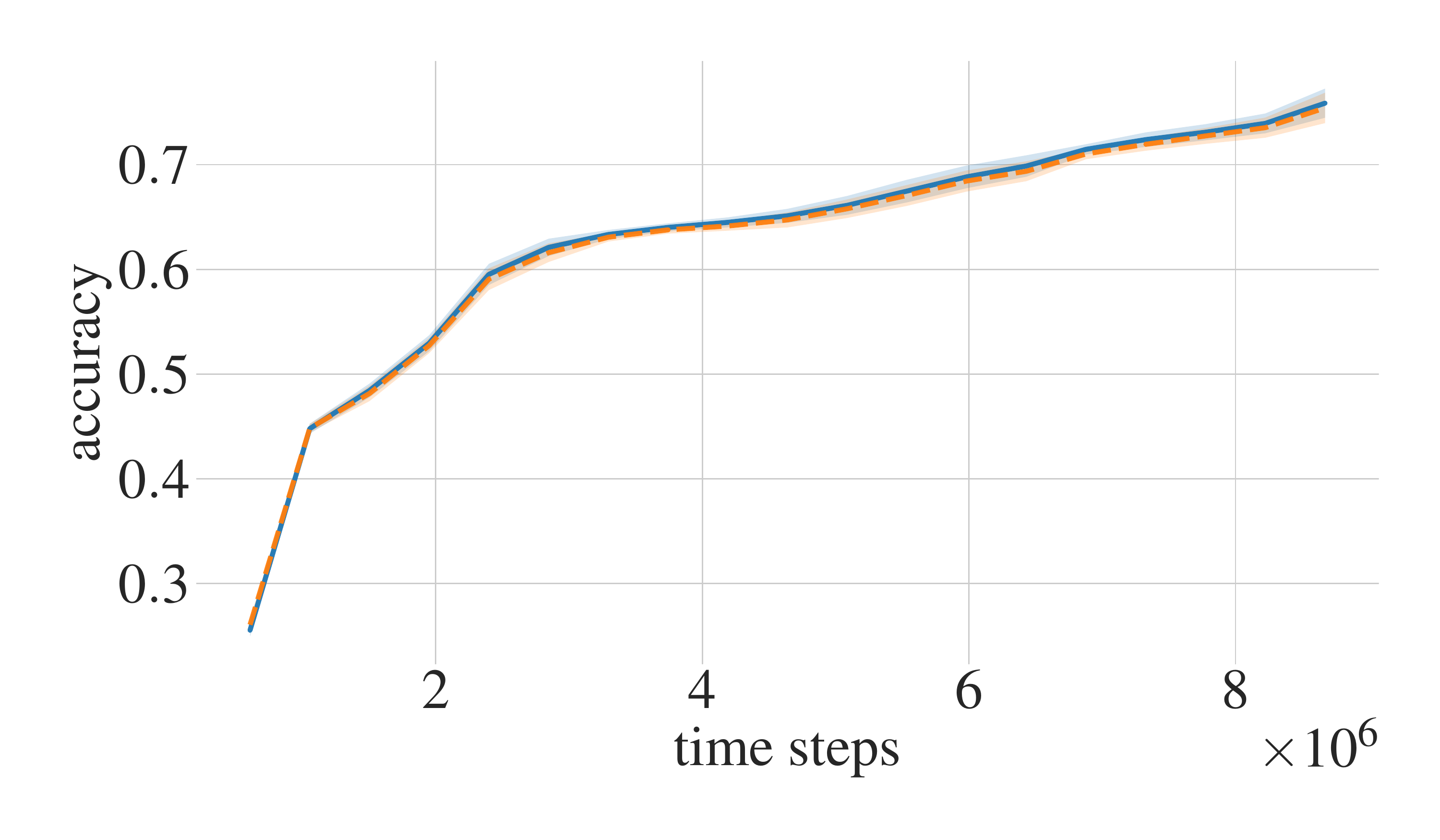}
    \label{fig:accuracy_color_shape}
  }
  \caption{We visualize both the training- and validation accuracy of our \hipss{} model in all task modes. In the first three modes (\autoref{fig:accuracy_default}, \autoref{fig:accuracy_default}, and \autoref{fig:accuracy_shape}), the seq2seq exceeds the word-level accuracy of 80\%. However, in the most demanding setup (\autoref{fig:accuracy_color_shape}), the accuracy is distinctly lower due to the larger number of possible instructions and combinations of object properties (cf. \autoref{tab:task_modes}).}
  \label{fig:accuracy}
\end{figure}

While \heir{} is faster than \hipss{} in the early stages of training (cf. \autoref{fig:success_color} and \autoref{fig:success_color_shape}), \hipss{} outperforms it in almost all cases later on.
A possible explanation for the superior performance of the predicted instructions of \hipss{} over the synthetic instructions of \heir{} is that the predicted instructions are more diverse than the expert instructions, and sometimes even incorrect.
Therefore, they might alleviate the issue of hindsight bias, which Bai et al. \cite{Bai_AddressingHindsight_2021} formulate as difference between the likelihoods of collected episodes conditioned on different goals.
In other words, we assume that the policy generates a similar trajectory under the hindsight instruction $\hat{g}_{\ell}$, which is not necessarily true because a different instruction $g_{\ell}$ was initially given.
More precisely, the policy understood the instruction correctly, but performed the wrong action-selection, instead of misunderstanding the instruction but performing correctly.
In the former case, providing a hindsight instruction and a sparse reward perturbs the learning greatly (cf. \heir{} in \autoref{fig:success_color_shape}).
An additional explanation of the performance difference is that the validation accuracy of our \hipss{} model is not at 100\%, but in the range of $70-85\%$ (depending on the task mode, see \autoref{fig:accuracy}).
This adds noise to the training and improves the policy's robustness, as also observed in related research \cite{Haarnoja_SoftActorCritic_2018,Cideron_HIGhERImproving_2020}.
In \autoref{fig:accuracy}, we underpin this claim by plotting the training and validation accuracy of our \hipss{} model for each task mode.
We employ a 1 to 5 ratio to store samples inside the training- and validation dataset.
This decision is made at the end of each successful episode (cf. \autoref{fig:hindsight_learning}).

\section{Conclusion}%
\label{sec:conclusion}

In this paper, we present three methods to improve the sample-efficiency in language-conditioned reinforcement learning for robotics.
Our first contribution is \heir{}, a mechanism for hindsight expert instruction replay.
Secondly, \hipss{}, a seq2seq model for hindsight instruction prediction, trained with trajectories collected during training only.
And finally, we present a class of language-focused tasks implemented with synthetic expert capabilities as part of our environment \envname{}.
Our results indicate that the self-predicted instructions provide the best performance gain, especially for difficult tasks.
In this article, we tested four different command modes. To further underpin our research, we plan to extend our work to real robots, e.g., our humanoid NICO robot \cite{Kerzel_NICO_2017,Eppe_CombiningDeep_2017} and a more diverse collection of simulated environments.
Furthermore, we consider a joint architecture to combine the benefits of both approaches as future work, where one uses \heir{} to accelerate the early training stages, followed by hindsight replay with \hipss{}.

\bibliographystyle{IEEEtran}
\bibliography{main}

\end{document}